\documentclass[a4paper,fleqn]{cas-dc}

\usepackage[authoryear]{natbib}
\usepackage{framed,multirow}
\usepackage{stfloats}
\usepackage{rotating}
\usepackage{graphicx}
\usepackage{textcomp}
\usepackage{amsmath,amsfonts,amssymb}
\usepackage{setspace}
\usepackage{tocloft}
\usepackage{colortbl}
\usepackage[dvipsnames]{xcolor}
\usepackage{pgfplots}
\usepackage{makecell}
\usepackage{latexsym}
\usepackage{url}
\usepackage{hyperref}

\definecolor{newcolor}{rgb}{.8,.349,.1}
\definecolor{LightCyan}{rgb}{0.88,1,1}
\definecolor{thistle} {rgb}{0.84,0.74,0.74}
\definecolor{lemon}{rgb}{1,0.97,0.8}
\definecolor{lavender}{rgb}{0.94,0.83,0.95}
\definecolor{aqua}{rgb}{0.49,1,0.83}
\definecolor{pink}{rgb}{1,0.65,0.9}
\definecolor{gray}{rgb}{0.87,0.87,0.87}
\definecolor{newcolor}{rgb}{.8,.349,.1}

\begin{document}

\shorttitle{Multimodal Deep Learning RCC Prognosis via CT and Clinical Data}    

\shortauthors{Mahootiha et al.}  

\title [mode = title]{Multimodal Deep Learning for Personalized Renal Cell Carcinoma Prognosis: Integrating CT Imaging and Clinical Data}  

\author[1,2]{Maryamalsadat Mahootiha}
\cormark[1]
\ead{marymaho@uio.no}
\cortext[1]{Maryamalsadat Mahootiha}

\author[2]{Hemin Ali Qadir}

\author[2]{Jacob Bergsland}

\author[2,3]{Ilangko Balasingham}

\affiliation[1]{organization={The Intervention Centre},
            addressline={Oslo University Hospital}, 
            city={Oslo},
            postcode={0372},
            country={Norway}}
\affiliation[2]{organization={Faculty of Medicine},
            addressline={University of Oslo}, 
            city={Oslo},
            postcode={0372},
            country={Norway}}
\affiliation[3]{organization={Department of Electronic Systems},
            addressline={Norwegian University of Science and Technology}, 
            city={Trondheim},
            country={Norway}}            

\begin{abstract}
\\\textbf{Background and Objective}: Renal cell carcinoma represents a significant global health challenge characterized by a low survival rate. The aim of this research was to devise a comprehensive deep-learning model capable of predicting survival probabilities in patients with renal cell carcinoma by integrating CT imaging and clinical data and addressing the limitations observed in prior studies. The aim is to facilitate the identification of patients requiring urgent treatment.\\
\textbf{Methods}: The proposed framework comprises three modules: a 3D image feature extractor, clinical variable selection, and survival prediction. The feature extractor module, based on the 3D CNN architecture, predicts the ISUP grade of renal cell carcinoma tumors linked to mortality rates from CT images. A selection of clinical variables is systematically chosen using the Spearman score and random forest importance score as criteria. A deep learning based network, trained with discrete LogisticHazard-based loss, performs the survival prediction. Nine distinct experiments are performed, with varying numbers of clinical variables determined by different thresholds of the Spearman and importance scores.\\
\textbf{Results}: Our findings demonstrate that the proposed strategy surpasses the current literature on renal cancer prognosis based on CT scans and clinical factors. The best-performing experiment yielded a concordance index of 0.84 and an area under the curve value of 0.8 on the test cohort, which suggests strong predictive power.\\
\textbf{Conclusions}: The multimodal deep-learning approach developed in this study shows promising results in estimating survival probabilities for renal cell carcinoma patients using CT imaging and clinical data. This may have potential implications in identifying patients who require urgent treatment, potentially improving patient outcomes. The code created for this project is available for the public on: \href{https://github.com/Balasingham-AI-Group/Survival_CTplusClinical}{GitHub}
\end{abstract}


\maketitle

\section{Introduction}
\subsection{Overview}
Renal cell carcinoma (RCC) is a prevalent malignancy in adults and constitutes around 90\% of all kidney tumors \citep{rccmortality1}. RCC develops in the tubules that filter blood and produce urine in the kidney \citep{rccmortality1}. If not detected and treated early, RCC can metastasize to other organs, such as lungs and bones, and become life-threatening \citep{sung2021global}. The global incidence of RCC has been rising, which may be attributable to the easy availability of more improved diagnostic modalities, greater use of medical imaging, and changes in lifestyle factors \citep{cancerstatistic, rccmortality2}. Treating RCC early is crucial for improving patient outcomes and enhancing both survival rates and quality of life \citep{rccmortality2}.

Survival analysis is a statistical technique used to investigate the time duration until a critical event occurs, such as death or disease recurrence, and is widely used in oncology. The analysis involves examining time-to-event data to estimate the probability of an event occurring over a specified period while accounting for censoring. This statistical technique allows for the inclusion of individuals who did not experience the event of interest by the end of the study period \citep{survanalysis}. 

Survival analysis is vital for RCC patients as it informs treatment decisions and enables clinicians to determine the optimal course of action, including therapy type, the intensity of treatment, and the need for palliative care or supportive measures \citep{importancesurvan}. Radiological data is essential for cancer survival analysis and prognosis, revealing tumor features, heterogeneity, therapy planning, and response evaluation. Clinicians can use this data to improve patient outcomes and survival prospects \citep{radiologicalinprognosis}. Clinical experts may make erroneous predictions or misinterpret medical images, which can result in incorrect prognosis and treatment decisions. In fact, approximately 20 million radiology reports are estimated to contain clinically significant errors annually \citep{errorbydoctors}. Furthermore, there may be a shortage of expert radiologists in certain regions or healthcare settings. Therefore, the implementation of artificial intelligence (AI) technologies can potentially aid in addressing these issues \citep{AIcliniciancomparison}. 

AI has the potential to improve the accuracy and efficiency of medical image analysis, particularly through the utilization of convolutional neural networks (CNN), which can capture patterns and features that may not be easily detectable by human observers \citep{AIinmedicalimaging1}. These algorithms can analyze large amounts of data quickly and accurately, reducing the potential for human error and improving diagnostic accuracy \citep{AIinmedicalimaging2}. The use of AI in survival analysis has also shown promise since it has the potential to enhance the precision of prognostic models and facilitate personalized treatment \citep{MLinsurvanalysis}. 

This study seeks to devise a multimodal AI-driven algorithm capable of predicting personalized survival probabilities utilizing CT images and clinical data, addressing challenges such as potential inaccuracies by clinicians and the scarcity of experts in radiological image interpretation. Our objective is to utilize a multimodal survival analysis strategy to achieve enhanced precision in forecasting survival probabilities. To investigate this, we classify RCC tumors in CT images according to the International Society of Urological Pathology (ISUP) grading \citep{ISUP} system. This system serves as a means to evaluate cancer severity by examining the morphological characteristics of tumor cells under microscopic observation, and it is closely associated with mortality rates \citep{ISUP2}. After the classification process, radiomic features are extracted and subsequently incorporated as input factors within our proposed survival model. Additional inputs encompass pertinent clinical variables pertaining to individual patients. By integrating radiomic features and clinical variables, we endeavor to estimate survival probabilities employing a methodology that is non-linear and non-proportional, offering a more robust, realistic, and accurate survival estimation.

\subsection{Related Work} 
\label{relatedwork}
In statistics, the Cox proportional hazards (CPH) model \citep{CPH} is the gold standard for modeling survival analysis using censored observations. CPH is limited by its linear nature, which fails to capture non-linear relationships between input data and the risk of an event occurring, e.g., death. However, the advent of AI and deep learning (DL) has opened new avenues for modeling survival analysis, allowing for the exploration of complex, non-linear relationships. DL-based models, such as Cox-nnet \citep{cox-nnet} and DeepSurv \citep{DeepSurv}, have been developed to address the limitations of the CPH model and enable the identification of novel prognostic factors. But they still face a fundamental constraint imposed by the proportional hazards assumption of CPH. CPH assumes that the effect of a patient's covariates on the risk of death remains constant over time, resulting in proportional predictions for all patients. However, this assumption may not be reflective of the true clinical situation, leading to survival curves that do not intersect.

Recent developments in statistical modeling have led to innovative solutions to address the limitations of the CPH model in survival analysis. Two important methods that have been proposed to address the linearity and proportionality constraints of CPH are multivariate time-to-event logistic regression (MTLR) \citep{MTLR}, and Nnet-survival \citep{Nnetsurvival}. MTLR is a method that extends logistic regression to time-to-event data by modeling the joint probability of multiple events. This approach allows for the incorporation of time-dependent covariates and can handle non-proportional hazards, making it a valuable tool for survival analysis. Nnet-survival, on the other hand, involves calculating the discrete conditional hazard rate at each time period. This concept has been established for several decades \citep{firstmodel_before_nnet} and was recently applied to contemporary DL approaches, leading to the development of Nnet-survival. This approach makes it possible to have non-proportional hazard probability curves for different patients.

Multimodal DL \citep{mutimodal_DL}, a framework that leverages DL techniques to learn from multiple data modalities, including tabular, images, and audio, can be particularly useful in medical applications. With the availability of diverse data types such as clinical information, radiological images, and medication records, the application of multimodal DL can help capture complex relationships between the model inputs and outputs. 

Previous studies have employed various approaches to conduct survival analysis, focusing on using radiological images or integrating radiological images with clinical variables to enhance survival estimation. \citet{lungnet:surv} developed a shallow CNN in conjunction with Cox loss to predict the prognosis of lung cancer patients using computed tomography (CT) image data alone. \citet{CTOvarian:surv} presented a CNN autoencoder-based survival model incorporating Cox loss for predicting recurrence in patients with high-grade serous ovarian cancer, relying solely on CT scans. \citet{DeepMMSA:surv} developed a regression-based survival model for non-small cell lung cancer patients, effectively integrating imaging and clinical data to enhance the accuracy of survival predictions by employing the mean squared error (MSE) loss function. \citet{gastric:surv} introduced a risk prediction model for assessing overall survival in gastric cancer patients, incorporating both CT images and clinical variables as inputs and utilizing a specialized loss function. \citet{nasopha:surv} presented a CNN-based model using Cox survival loss to predict survival outcomes in patients diagnosed with stage T3N1M0 nasopharyngeal carcinoma using magnetic resonance (MR) imaging and clinical variables. Lastly, \citet{paper1compare} explored the potential of radiomic features and clinical variables in predicting the survival group of lung cancer patients. The authors employed image analysis techniques, rather than DL methods, to extract radiomic features, and utilized a random forest classifier.

\subsection{Our Contributions}
This study differs from the previous study by presenting a novel multimodal approach to predicting nonlinear and non-proportional survival curves for patients afflicted with RCC by employing both CT images and clinical data. Moreover, our study is distinguished as the first to systematically explore the impact of varying combinations of clinical variables and CT images on survival prediction performance, thereby shedding light on the importance of selecting appropriate data sources for accurate survival estimations.

Our proposed survival model offers several notable advantages over previous studies, which can be delineated in the following manner: 1) By incorporating 3D inputs and 3D convolutional layers, our model retains comprehensive information from the data, mitigating any potential loss of critical details pertaining to the interface between tumor and healthy tissue. 2) Our methodology enables the forecasting of non-proportional survival analyses, producing outcomes that are more relevant to clinical situations. 3) In comparison to previously reported literature, our survival model demonstrates superior performance indices, highlighting its efficacy. 4) A key feature of the proposed model is its ability to generate individualized survival curves for each patient, allowing for a more personalized assessment. 5) To elucidate the nuances of survival model performance, we conduct an analysis of varying combinations of clinical variables and CT images, providing valuable insights into the optimization of survival estimation. 6) In addition to conventional metrics for evaluating survival models, we also employ the violin diagram to visualize the distribution of survival probabilities in our survival model's outputs.

\section{Methods}
\begin{figure*}[t]
    \centering
    \vspace{1ex}
    \includegraphics[scale=0.265]{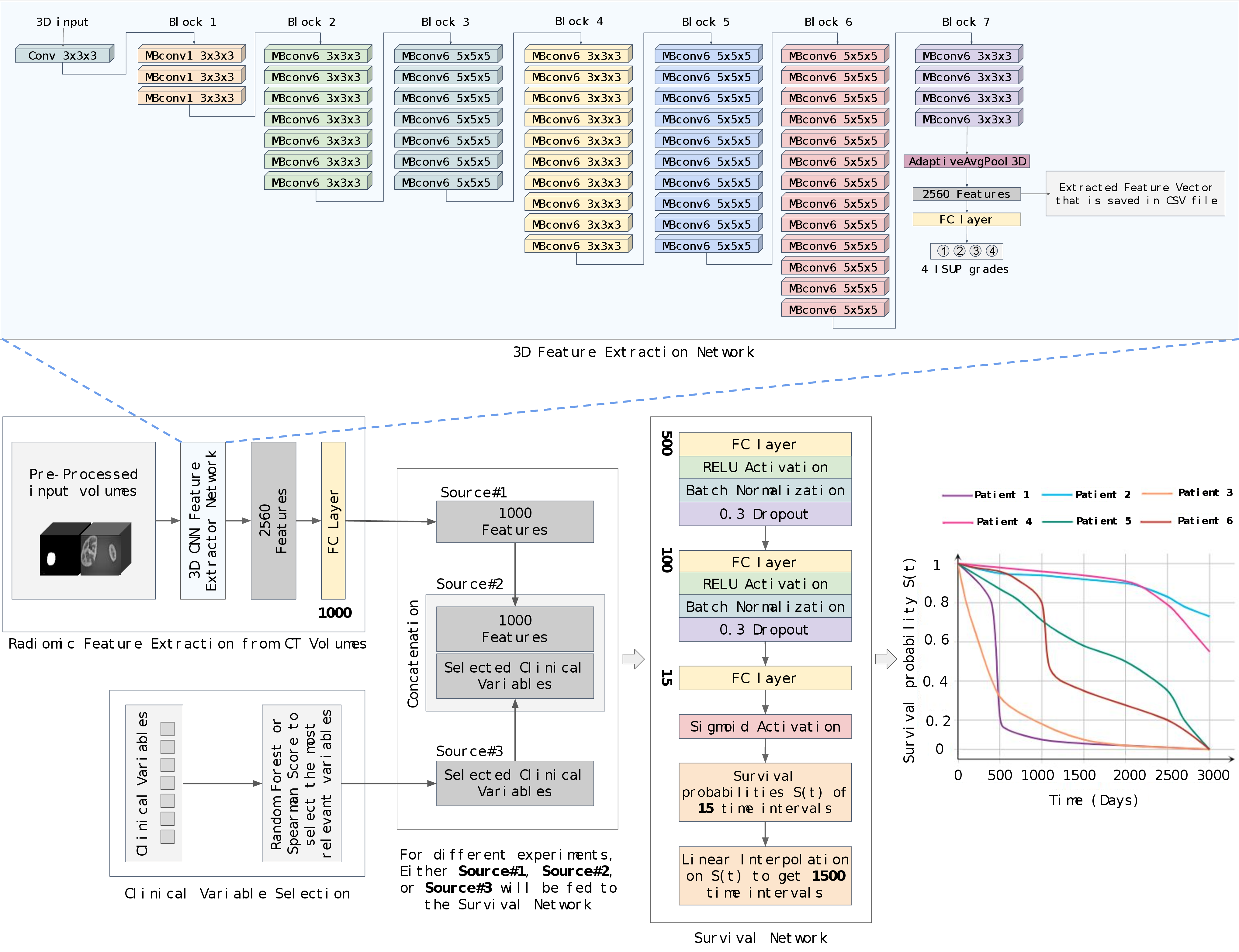}
    \caption{\footnotesize{The comprehensive framework presented herein is composed of three primary modules. Module 1 encompasses feature extraction from CT volumes, wherein features are derived through the classification of CT images based on ISUP grades. Subsequently, a fully connected layer consisting of 1000 neurons is employed to reduce the radiomic feature size from 2560. Module 2 focuses on the judicious selection of clinical variables, which are merged with CT image features utilizing the Spearman correlation score and random forest importance score. Module 3 pertains to the survival network, which accepts input from three sources: CT image features, clinical variables, and a combination of both. The survival network's output consists of survival probabilities for 15 discrete time intervals, which are subsequently converted into 1500 time points through interpolation. This process facilitates the visualization of continuous survival curves for individual patients.}}
    \label{The_Framework}
\end{figure*}

Fig. \ref{The_Framework} illustrates our entire approach for modeling survival analysis. It takes as inputs two data modalities: 1) CT volumes and 2) clinical variables. Motivated by the success of CNNs in image analysis and cancer prognosis, we present a CNN-based architecture for CT image feature extraction relevant to prognosis in our methodology. We utilize 3D CNNs to extract features from the three dimensions within the tumor volume motivated by \citet{3dnetlungclassification}. Subsequently, we integrate clinical information with the CT image features for survival analysis. Our method comprises three modules: (1) CT image feature extraction, (2) clinical variables selection, and (3) survival prediction. Within the scope of our scholarly investigation, the feature extractor network and the survival network are subjected to independent training processes as opposed to being trained concurrently.

\subsection{Radiomic Feature Extraction from CT Volumes}\label{featureextraction}
We suggest classifying RCC tumors in CT images into ISUP grades (1, 2, 3, and 4) to obtain radiomic features relevant to prognosis. The CT volumes go through a 3D CNN feature extractor network to pull out these features. After that, we can integrate the clinical variables with the extracted radiomic features. We choose ISUP grade for classification as it has been shown to have a strong correlation to tumor recurrence, metastasis, and mortality \citep{ISUP1}. Higher ISUP grades are indicative of a worse prognosis and higher mortality rate, whereas lower grades are associated with a better prognosis, and lower mortality rate \citep{ISUP3}.

For the feature extractor network, the classifier, in our study, we select EfficientNet \citep{effnet}, which is a state-of-the-art CNN architecture developed by Google researchers for image classification. This architecture employs the compound coefficient method to scale up models efficiently. The largest model, EfficientNet B7, achieved the best performance compared to other variants. The EfficientNet layers utilize MBConv \citep{Mobileconv}, a type of convolutional block that can capture complex features in images while using fewer parameters and less computation compared to traditional convolutional blocks.

To accommodate three-dimensional (3D) image data such as CT volumes, we adapt the exact architecture of EfficientNet B7 and transform it into a 3D CNN model. By doing so, features are extracted in all three-dimensional directions within the tumor volume, taking the third-dimensional spatial information into account. Hence, the employed feature extraction network operates in a three-dimensional (3D) domain, wherein the input comprises image volumes that have undergone preprocessing and concatenated with the annotations of tumor segmentations. This network classifies the RCC tumors into four ISUP grades. Our group has undertaken a separate, comprehensive study focused on the classification of RCC according to ISUP grading systems \citep{maryam2012isup}. The architecture encompasses a combination of convolutional layers, MBConv layers, an Adaptive Average Pooling layer, and a series of fully connected (FC) layers, respectively. 

The Adaptive Average Pooling layer, which acts as a bridge between CNN and FC layers, can be used for feature extraction. This layer reduces the number of parameters and computational complexity required for classification while preserving crucial information about image features \citep{imagenet_averagepolling}. We extract the outputs from the Adaptive Average Pooling layer to create feature vectors for every patient. Subsequently, the output of the Adaptive Average Pooling layer is flattened, and the resulting image features are converted to feature vectors. The initial feature vector dimension is 2560, and our objective is to reduce it to 1000 to streamline integration with clinical variables. We attempt to achieve this reduction by employing an FC layer with 2560 input features and 1000 output features. These vectors are then saved as a CSV file for feeding to the survival network. Following the feature extraction and storage in a CSV file, normalization is performed to standardize the data based on the mean and standard deviation.

\subsection{Clinical Variables Selection}\label{clinicalselection}
Our objective is not to incorporate all clinical variables with CT image features for the purpose of survival prediction. Rather, we intend to explore the feasibility of using a smaller subset of variables (those that are more relevant to prognosis) in conjunction with CT image features to achieve improved results in survival prediction. To this end, we aim to evaluate various combinations of clinical variables. In order to identify the most relevant clinical variables for predicting survival times, we employ two well-established methods: (1) the Spearman correlation score \citep{spearman} and (2) the importance score of a random forest regressor \citep{RFimportancescore}. These approaches help us to identify the most informative clinical variables to include in our survival model and achieve more accurate predictions of survival outcomes for each patient.

Spearman's rank correlation coefficient is a non-parametric statistical method that quantifies the strength and direction of the relationship between two variables. It is primarily used to assess the existence of a monotonic association between two variables and is less sensitive to non-linear relationships and non-normal distributions compared to the parametric Pearson correlation coefficient. We calculate the Spearman correlation coefficients between clinical variables and survival times, forming a correlation matrix. This matrix represents the pairwise correlation coefficients between each clinical variable and survival times. The correlation coefficient values vary from -1 to 1, with -1 representing a strong negative connection, 1 representing a strong positive connection, and 0 representing no correlation. 

On the other hand, random forest regression can generate an importance score for each predictor variable.
To acquire importance scores, we develop a random forest model consisting of 100 decision trees to estimate survival times based on clinical variables. The importance scores originate from the average decrease in the model's prediction error due to each feature, considering all the random forest's decision trees. Subsequently, the clinical variables are ranked according to their importance scores to discern the most influential variables for survival time prediction. Higher importance scores signify a more substantial impact of a variable on the model's predictive performance.

\subsection{Modeling Survival Estimation}
In this subsection, we focus on the critical aspects of modeling survival estimation, an essential component of our proposed method. We have organized this subsection into three parts: 1) survival network, where we describe the architecture and design choices for the survival network, which is responsible for estimating survival probabilities; 2) input to the survival network, which details the features and data used as input to the network, such as clinical variables and radiomic features extracted from the 3D CNN feature extractor; 3) loss function for modeling survival estimation, in which we discuss the choice of loss function employed to optimize the survival network.
\subsubsection{Survival Network}
The survival network, shown in Fig. \ref{The_Framework}, consists  of three FC layers, comprising 500, 100, and 15 neurons, respectively. As our model is a discrete-time survival model, the final layer contains 15 neurons representing survival probabilities for 15 distinct time intervals. The network utilizes a rectified linear unit (ReLU) activation function in the intermediate layers and a sigmoid activation function in the last layer. In an effort to enhance the generalization capabilities of the model, a dropout rate of 0.3 is incorporated, accompanied by the implementation of batch normalization subsequent to the initial two FC layers. Subsequently, linear interpolation with 100 points is employed to transform the outputs into a set of 1500 values, enabling the generation of continuous survival curves for patients. We achieve the optimal architecture through a grid search of hyperparameters to find the best evaluation metrics for survival analysis. 

\subsubsection{Input to the Survival Network}
The inputs to the survival network are derived from one of three sources: CT image features, clinical variables, or a combination of CT image features and clinical variables. In this study, we do a series of nine experiments, each using one of these three sources for survival prediction. In Section \ref{exps}, a full explanation of these experiments will be given.

\subsubsection{Loss Function for Modeling Survival Estimation}\label{lossfunction}
We adapt our survival model loss function based on discrete logistic hazards similar to the loss used in Nnet survival \citep{Nnetsurvival} to predict survival probabilities over M days (weeks, months, or years) which M is the maximum follow-up period. In order to employ the discretized hazard function, it is essential to convert continuous survival times into discrete intervals. To achieve this, a judicious selection of appropriate time intervals is undertaken to discretize the continuous survival times, with the preferred choice being equidistant intervals. Subsequently, each observed survival time is allocated to its respective time interval, effectively transforming the continuous data into a discrete format. We developed a loss function that used  a vectorized form of likelihoods for censored and uncensored patients. The loss function is given by:
$$
\operatorname{L}=\mathrm{-}\sum_{x=1}^{p}\sum_{i=1}^{n}\left(\begin{array}{l}
\ln \left(1+\operatorname{surv}_{s}(x)(i) \cdot\left(\operatorname{surv}_{\text {pred }}(x)(i)-1\right)\right) \\
+\ln \left(1-\operatorname{surv}_{f}(x)(i) \cdot \operatorname{surv}_{\text {pred }}(x)(i)\right)
\end{array}\right),
$$
where $p$ denotes the number of patients in a batch, and $n$ represents the number of discrete time intervals (15). $\operatorname{surv_{pred}}(x)(i)$ signifies the predicted outcome of the survival model for patient $x$ at time interval $i$, which can be either 0 for a patient who died during interval $i$ or 1 for a patient who remained alive in interval $i$. Each patient's death or censoring time, $t$, is determined based on the ground truth survival time given in a dataset. The ground truth vectors $\operatorname{surv_{s}}$ and $\operatorname{surv_{f}}$ for the survival model are of length $n$ for every patient. Vector $\operatorname{surv_{s}}$ corresponds to the time intervals when the patient survived, while vector $\operatorname{surv_{f}}$ denotes the specific time interval when the death occurred. For uncensored patients in the time interval $i$:
$$
\begin{aligned}
&\operatorname{surv}_{s}(x)(i)= \begin{cases}1, & \text { if } t_{x} \geq t_{i} \\
0, & \text { otherwise }\end{cases} \\
&\operatorname{surv}_{f}(x)(i)= \begin{cases}1, & \text { if } t_{i-1} \leq t_{x}<t_{i} \\
0, & \text { otherwise }\end{cases}
\end{aligned}
$$

for censored patients in the time interval $i$:
$$
\operatorname{surv}_{s}(x)(i)= \begin{cases}1, & \text { if } t_{x} \geq \frac{1}{2}\left(t_{i-1}+t_{i}\right) \\ 0, & \text { otherwise }\end{cases}
$$
and
$$
\operatorname{surv}_{f}(x)(i)=0.
$$

The dot product within the loss function assesses the similarities between the predicted vector and the ground truth vector. We trained the survival networks with the help of pycox v0.2.0.3 library \footnote{https://github.com/havakv/pycox}.

\section{Experimental Setup}
In this section, we describe the experimental setup employed in our study, which is divided into four main parts: experimental dataset, training the 3D CNN feature extractor network, training the survival network, and the experiments conducted. First, we present the datasets used in our study and discuss their characteristics, source, and any preprocessing steps undertaken. Next, we outline the process of training the 3D CNN feature extractor network, followed by the training of the survival network. Finally, we describe the experiments conducted. A comprehensive experimental setup ensures the reproducibility of our results and allows for a fair comparison with other studies in the field.
\subsection{Experimental Dataset}
The selection of appropriate datasets and their preparation plays a crucial role in the evaluation of our proposed method. In this subsection, we provide an overview of the dataset used in our experiments and the steps taken to prepare the data for our study. We have divided this subsection into three parts: the KiTS21 dataset, dataset splitting, and clinical data preparation. First, we discuss the KiTS21 dataset, its characteristics, and its source. Next, we describe the dataset-splitting process, explaining the rationale behind the chosen method and the proportions used for training, validation, and testing. Finally, we detail the clinical data preparation, including any necessary preprocessing and data normalization procedures.
\subsubsection{KiTS21 Dataset} We used the {KiTS21} \citep{kits2} dataset to train and test our proposed framework. The dataset comprises 300 patients who underwent either partial or complete nephrectomy for suspected kidney cancer between 2010 and 2020 at the M Health Fairview or Cleveland Clinic medical facility and includes both clinical data and CT scans with manually annotated kidneys and tumors (ground-truth labels). The primary objective of collecting this dataset was to apply segmentation algorithms.

We selected this dataset for its comprehensive clinical information, precise annotations, and ample subject numbers. The dataset contains three files, including CT scan volumes (NIFTI format), annotation volumes (NIFTI format), and clinical data (JSON format). The annotation volumes consist of manual segmentations of the kidneys, tumor(s), and cyst(s). In this study, we used 41 clinical variables from this JSON file. All critical clinical information, such as pathology results, is included in this file \citep{kits1}. Notably, this data was originally obtained from the Cancer Imaging Archive in DICOM format, while the clinical data was provided in a single CSV file.

\subsubsection{Dataset Splitting} 
To train the classify network that can be used as the radiomic feature extraction for survival prediction, we excluded 56 patients with empty ISUP grade values from the original dataset. The remaining dataset contained 244 patients, of which 32 had dead events and 212 had censored time. The maximum observation time was 3000 days (which refers to the M variable in Section \ref{lossfunction}), and the median observation time was 644 days. We performed three-fold cross-validation for the ISUP grading classification to create three different subsets for training, validation, and testing. The division of the dataset into three folds was based on the number of deceased and censored patients to ensure that each subset contained the same proportion of deceased individuals. Each fold included 57\% of the total dataset for training, 10\% for validation, and 33\% for testing. The training subset had 10\% of patients who died, the validation subset had 33\%, and the test subset had 13\%. After dividing the dataset into three folds, we increased the number of samples in each train and validation subset by doing multiple augmentations (discussed in \ref{preprocessct}). 

The optimal fold for the classification model was determined based on the F1-score, as delineated in \ref{best_fold}. This selected fold was subsequently employed for training, validation, and testing within the survival network, excluding the utilization of augmented samples. Two distinct networks were employed for ISUP grade classification and survival analysis; however, they were trained using identical subjects within the training, validation, and test datasets. This approach was adapted to preclude the introduction of the classification network's training data as the validation or test dataset for the survival analysis network, thereby avoiding the overestimation of the survival analysis network's performance due to heightened accuracy in detecting ISUP grades within the training dataset.

\subsubsection{Clinical Data Preparation}
The clinical data used in training the survival network consisted of 38 variables classified into two categories: continuous numerical and categorical. In order to facilitate their usage in the survival model, the categorical variables were transformed into discrete numerical values, such as gender. In contrast, the continuous numerical variables, such as pathologic size, were normalized based on the mean and standard deviation to facilitate effective interpretation by the survival model.

\subsection{Training the 3D CNN Feature Extractor}
In this subsection, we elaborate on the process of training the 3D CNN feature extractor, a critical component in our proposed method. This subsection is divided into three parts: 1) preprocessing of CT image volumes, which is a necessary step before training the 3D CNN feature extractor to guarantee consistent input data and enhance the network's performance; 2) training details of the classifier, encompassing aspects such as the chosen loss function, number of epochs, optimizer, and learning rate; 3) best fold selection for radiomic feature extraction, a crucial step following the training of the 3D CNN feature extractor, which involves selecting the optimal fold to ensure the highest quality features for the subsequent survival network.
\subsubsection{Preprocessing of CT Image Volumes}\label{preprocessct}
Before commencing the preprocessing phase for CT volumes, image augmentations were implemented as a strategy to address the inherent imbalance in the dataset, as well as the paucity of training samples. A combination of positional augmentations, such as flipping, rotation, and affine transformations, along with noise augmentations, including Gaussian noise, Gibbs noise, and space spike noise, were employed to enhance the diversity and generalizability of the dataset. Before the ISUP grade classification, image preprocessing is applied to improve the quality of the input images and their radiomic features for better interpretation of the input \citep{torchio,preprocess1}. As recommended in the MIT challenge\footnote{\href{http://6.869.csail.mit.edu/fa17/miniplaces.html}{http://6.869.csail.mit.edu/fa17/miniplaces.html}}, all volumes were resized to $128\times128\times128$. We also resampled the volumes based on one millimeter isotropic voxel size, which has been recommended as a standard voxel size by previous studies in medical imaging \citep{medicalimagesegment,brainimageclassification}. Additionally, all volumes were reoriented to the RAS (Right, Anterior, and Superior) orientation, which is the most commonly used orientation in medical images \citep{medicalimagesegment,brainimageclassification,DLmedicalimage}. We utilized intensity normalization based on the Z-score in medical imaging \citep{torchio,Intensitybias}. For kidney image and tumor segmentation, identical image preprocessing steps were employed, with the exception that intensity normalization was not applied for tumor segmentation.

As part of our image preprocessing pipeline, we employed a concatenation step to enhance the performance of our 3D EfficientNet-B7 model in identifying kidney tumors. Specifically, we combined the extracted kidney images with their corresponding manual tumor segmentations to enable the model to focus on the surface patterns of the tumors \citep{preprocess1}. This image concatenation approach serves to enrich the input volume with additional information pertaining to the location and size of the tumors. If the model were to be trained solely on the kidney images without the inclusion of tumor location data, it could potentially pick up on irrelevant features and perform poorly on previously unseen data. Thus, the concatenation step helps to improve the model's generalizability and overall accuracy.

\subsubsection{Training Details}
To validate the robustness of the radiomic feature extractor network, we conducted three-fold cross-validation with three distinct train, validation, and test subsets, while maintaining the same hyperparameters for each training iteration. For training the 3D CNN feature extractor, we used the ADAM optimizer \citep{adamoptimizer} with a fixed learning rate of $1\times10^\mathrm{-4}$, and 50 epochs were run to optimize the network parameters. In addition, we employed the Cross-Entropy loss given by:
\begin{equation}
    \label{crossentopyloss}
   L=\mathrm{-} \sum_{i=1}^{n} t_i\times log(p_i),
\end{equation}
where $t_i$ is the true ISUP class and $p_i$ is the softmax probability for the $ith$ class, and $n$ is the number of ISUP classes (4 in this study). The 3D feature extractor was trained using PyTorch v1.11.0 on a workstation equipped with an Nvidia GeForce RTX 3090 GPU, an AMD Ryzen 7 5800X 8-Core Processor, and 32 GB of RAM.

\subsubsection{Best Fold Selection for Radiomic Feature Extraction}
\label{best_fold}
We used precision, recall, and F-score in the evaluation of our feature extractor network, as these fundamental metrics are indispensable for assessing classification model performance. 

Precision, also known as the positive predictive value, quantifies the fraction of true positives out of the total instances predicted as positive by the model. Mathematically, precision can be defined as:
\begin{equation}
\text{Precision} = \frac{\text{TP}}{(\text{TP} + \text{FP})},
\end{equation} where TP denotes true positives and FP denotes false positives. 

Recall, alternatively referred to as sensitivity or true positive rate, measures the fraction of true positive instances among the total number of actual positive instances within the dataset. Recall can be mathematically represented as: \begin{equation}
\text{Recall} = \frac{\text{TP}}{(\text{TP} + \text{FN})},
\end{equation} where FN denotes false negatives.

The F-score, specifically the F1-score, constitutes the harmonic mean of precision and recall, delivering a single metric that balances both measures. The F1-score is particularly advantageous in situations with uneven class distributions, as it accounts for the trade-off between precision and recall. The F1-score can be calculated using the following equation: \begin{equation}
\text{F1-score} = 2 * \frac{(\text{Precision} * \text{Recall})}{(\text{Precision} + \text{Recall})}.
\end{equation} We calculated the average of four Precision, Recall, and F-scores that we gained for each ISUP class. We repeated this process three times for each of our three folds, giving us three average Precision, Recall, and F-scores. The second fold, with an average F-score of 0.84, was the best and selected as our final radiomic feature extractor that can be used the the input for the survival network.

\subsection{Training the Survival Network}
In the present study, we used a total of 500 epochs for training the survival network. To prevent overfitting and enhance generalization, early stopping was implemented with a patience level of 10. The model was optimized utilizing the Adam optimizer, accompanied by a learning rate of 0.01. The optimal learning rate selection was determined by applying the method put forth by Smith \citep{lrfinder}.

\subsection{Experiments}
\label{exps}
In our study to demonstrate the performance improvement of our proposed survival analysis framework, we conduct nine distinct experiments with different combinations of inputs. The first experiment involves solely CT image features, the second only involves clinical variables, and the third combines CT image features and clinical variables. The remaining six experiments are created by applying three distinct thresholds for each the Spearman correlation and the random forest regression importance score. The clinical variables are selected based on the thresholds in the last six experiments and then fed to the survival network. These experiments are then compared to each other to evaluate their effectiveness in predicting survival outcomes. Further details on the results of these experiments will be presented in Section \ref{results}.

\section{Results}
In this section, we present the evaluation of our survival model's performance, the experimental results, and a comparison with related previous studies. We have organized this section into four parts: 1) metrics for survival model performance evaluation, where we describe the evaluation metrics used to assess the performance of our proposed survival model; 2) experimental results from nine different experiments, in which we report and analyze the results obtained from a series of nine distinct experiments conducted to evaluate our method; 3) plotting violin diagram for survival distribution, which involves the visualization of survival distribution data using violin diagrams to provide a comprehensive understanding of the results; 4) discussion, where we compare our findings with those from related previous studies, highlighting the improvements and contributions made by our proposed method.
\subsection{Metrics for Performance Evaluation}\label{metricseval}
To assess the performance of our survival model, we used two key metrics: the time-dependent concordance index ($C^{td}$) and the cumulative dynamic area under the curve (AUC). $C^{td}$ extends Harrell's concordance index \citep{harrell_cindex}, a widely utilized measure for evaluating the discriminative power of survival models. The time-dependent C-index is specifically designed to address situations in which a model's predictive accuracy may vary over time. It gauges the model's capacity to accurately rank the predicted survival probabilities of subject pairs at a specific time point, taking censoring into account. The computation of $C^{td}(t)$ involves dividing the count of accurately ordered pairs by the total count of comparable pairs. The $C^{td}$ range between 0 and 1, where values approaching 1 signify superior predictive accuracy, while those nearing 0.5 indicate the model possesses no greater discriminative power than random chance. It has been established that the concordance index is excessively optimistic, particularly with an increasing number of censored patients in the dataset \citep{cindex_censor_problem}.

The cumulative dynamic AUC \citep{AUC_survival} extends the conventional AUC metric, a prominent measure for assessing binary classification models. This extension is tailored to specifically address censored data and time-varying predictions in the realm of survival analysis. Within this context, the cumulative dynamic AUC is computed for a designated time point t, quantifying the model's discriminatory capacity to distinguish subjects experiencing the event of interest by time t from those who do not. The cumulative dynamic AUC represents the area under the time-dependent Receiver Operating Characteristic (ROC) curve, which delineates the sensitivity (true positive rate) against 1-specificity (false positive rate) for different time points. Ranging from 0 to 1, the cumulative dynamic AUC reveals greater predictive accuracy as it approaches 1, while values nearing 0.5 indicate that the model's discriminatory power is no better than random chance.

In addition to standard metrics, we use violin plots, a novel approach, to observe survival model output distributions. This is the first study proposing the application of violin plots for the evaluation of survival models. High evaluation metrics may be misleading, as predicted survival probabilities may not match ground truth times of death. Violin plots serve as a valuable tool in visualizing model performance by exhibiting the distribution of predicted probabilities at the time of mortality for deceased individuals, as well as the distribution of predicted probabilities at the ultimate time point for censored subjects. For example, a distribution approximating zero for deceased patients signifies satisfactory model training, which consequently yields probability predictions in close proximity to zero.



\subsection{Experimental Results}
\label{results}
One of our study aims to investigate the impact of various combinations of clinical variables on the prediction of survival outcomes in patients with RCC. Specifically, we seek to identify the clinical features that contribute most significantly to the accurate prediction of patients' survival times. Initially, we conducted two independent analyses to evaluate the effectiveness of CT image features and clinical data individually with respect to their impact on the performance of our survival model. Subsequently, we explore the impact of merging CT image features with various combinations of selected clinical variables on the performance of the survival model. 

To this end, we developed nine distinct experiments (Exp). Table \ref{9experimentstable} shows the difference between these nine experiments in terms of their inputs and thresholds used for choosing the combination of clinical variables. Table \ref{9experimentstable} also reports the {C}-index and AUC obtained on the test subset from each experiment. We used the same survival network architecture in the nine experiments for a fair comparison. From experiment 4 to experiment 9, we applied different thresholds for the Spearman correlation score (S\_score) and random forest regression importance score (I\_score).

\begin{table}[h]
\centering
\caption{Differences of Experiments used for RCC survival analysis.}
\small
\begin{tabular}{|p{0.6cm}|p{2.7cm}|c|p{0.6cm}|p{0.5cm}|}
\hline
\textbf{Exp} & \textbf{Inputs} & \textbf{Theresholds} & \textbf{C-index} & \textbf{AUC}\\
\hline\hline

Exp1 & CT images Features & & 0.72 & 0.73\\
\hline\hline

Exp2 & 38 clinical variables & & 0.72 & 0.74 \\
\hline\hline

\multirow{2}{*}{Exp3} & CT images Features & & \multirow{2}{*}{0.82} &  \multirow{2}{*}{0.74}\\
\cline{2-3}
& 38 clinical variables & & &\\
\hline\hline

\multirow{2}{*}{Exp4} & CT images Features & & \multirow{2}{*}{0.79} &  \multirow{2}{*}{0.76}\\
\cline{2-3}
& 4 clinical variables & $\vert\text{S\_score}\vert \geqslant 0.1$ & & \\
\hline\hline

\multirow{2}{*}{Exp5} & CT images Features & & \multirow{2}{*}{0.83} &  \multirow{2}{*}{0.75}\\
\cline{2-3}
& 13 clinical variables & $\vert\text{S\_score}\vert \geqslant 0.05$ & & \\
\hline\hline

\multirow{2}{*}{Exp6} & CT images Features & & \multirow{2}{*}{0.81} &  \multirow{2}{*}{0.77}\\
\cline{2-3}
& 30 clinical variables & $\vert\text{S\_score}\vert \geqslant 0.01$ & &\\
\hline\hline

\multirow{2}{*}{Exp7} & CT images Features & & \multirow{2}{*}{0.77} &  \multirow{2}{*}{0.74}\\
\cline{2-3}
& 4 clinical variables & $\text{I\_score} \geqslant 0.1$ & &\\
\hline\hline

\multirow{2}{*}{\textbf{\textcolor{WildStrawberry}{Exp8}}} & \textbf{\textcolor{WildStrawberry}{CT images Features}} & & \multirow{2}{*}{\textbf{\textcolor{WildStrawberry}{0.84}}} &  \multirow{2}{*}{\textbf{\textcolor{WildStrawberry}{0.8}}}\\
\cline{2-3}
& \textbf{\textcolor{WildStrawberry}{17 clinical variables}} & $\text{\textbf{\textcolor{WildStrawberry}{I\_score}}} \textcolor{WildStrawberry}{\geqslant} \textbf{\textcolor{WildStrawberry}{0.01}}$ & &\\
\hline\hline

\multirow{2}{*}{Exp9} & CT images Features & & \multirow{2}{*}{0.84} &  \multirow{2}{*}{0.76}\\
\cline{2-3}
& 29 clinical variables & $\text{I\_score} \geqslant 0.001$ & &\\
\hline
\end{tabular}
\label{9experimentstable}
\end{table}

We adjusted three different thresholds for Spearman's correlation coefficient. As the threshold values decreased, we incorporated more clinical variables with weaker correlations to the patient survival time into the survival model. In contrast, we utilized three different thresholds for the importance score of the decision tree regressor. By lowering these threshold values, we gradually incorporated less important clinical variables in predicting survival times into the survival model.

According to Table \ref{9experimentstable}, the best evaluation metrics were obtained in experiment 8, in which the C-index and AUC are 0.84 and 0.8, respectively. The inputs to experiment 8 are the followings: CT images features, Localized Solid Tumor, Age at Nephrectomy, Congestive Heart Failure, Body Mass Index, Uncomplicated Diabetes Mellitus, Pathologic Size, Myocardial Infarction, Radiographic Size, Metastatic Solid Tumor, Hospitalization, Mild Liver Disease, Smoking History, Surgery Type, Gender, Tumor Histologic Subtype, Pathology T Stage, and Surgical Approach.

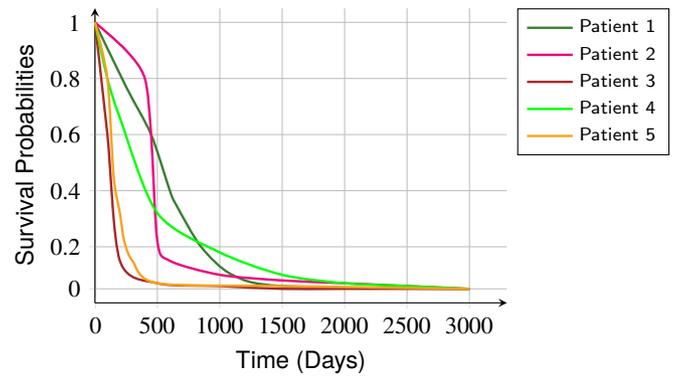
\begin{figure}[!h]
\centering
\begin{tikzpicture}  
\begin{axis} 
[domain = 0:100, 
axis on top=false, 
axis x line=middle, 
axis y line=middle, 
xlabel = \fontfamily{qhv}\small Time (Days), 
xlabel near ticks, 
ylabel = \fontfamily{qhv}\small Survival Probabilities,  
ylabel near ticks, 
width=7.0cm, 
legend style={font=\scriptsize, at={(1.025,1)},anchor=north west,legend columns=1}, 
height=5.5cm, 
clip  = true, 
xmin = 0,   
xmax = 3300, 
ymin = 0, 
ymax = 1.05, 
grid=major, 
ytick={0,0.05,0.2,0.4,0.6,0.8,1,1.2}, 
yticklabels={$0$,$0$,$0.2$,$0.4$,$0.6$,$0.8$,$1$}, 
xtick={0,5,500,1000,1500,2000,2500,3000,3500}, 
xticklabels={$0$,$0$,$500$,$1000$,$1500$,$2000$,$2500$,$3000$,$3500$},]  

\addplot [OliveGreen, line width = 0.9, smooth] coordinates {(0,1)(250,0.77)(450,0.6)(600,0.4)(650,0.35)(850,0.2)(1100,0.1)(1500,0.06)(3000,0.05)};  
\addlegendentry{Patient 1}

\addplot [RubineRed, line width = 0.9, smooth] coordinates {(0,1)(250,0.9)(400,0.8)(450,0.6)(500,0.22)(600,0.15)(1000,0.1)(1500,0.08)(2000,0.07)(3000,0.05)};  
\addlegendentry{Patient 2}

\addplot [Brown, line width = 0.9, smooth] coordinates {(0,1)(100,0.6)(200,0.15)(500,0.07)(1000,0.06)(1500,0.05)(2000,0.05)(3000,0.05)};  
\addlegendentry{Patient 3}

\addplot [green, line width = 0.9, smooth] coordinates {(0,1)(100,0.8)(200,0.66)(500,0.32)(1000,0.18)(1500,0.1)(2000,0.07)(2500,0.06)(3000,0.05)};  
\addlegendentry{Patient 4}

\addplot [YellowOrange, line width = 0.9, smooth] coordinates {(0,1)(100,0.8)(150,0.45)(200,0.32)(235,0.22)(300,0.15)(500,0.07)(1500,0.06)(3000,0.05)};  
\addlegendentry{Patient 5}
\end{axis}  
\end{tikzpicture} 
\caption{Survival Probabilities for five patients in the test cohort who died.}
\label{survivalcurve5patient}
\end{figure}

In order to evaluate the effectiveness of the survival model, ten unique individuals from the test cohort were selected, of which five had deceased from RCC, and five had censoring time to event. Subsequently, the survival curves for these patients were plotted, utilizing the survival probabilities derived from experiment 8. Fig. \ref{survivalcurve5patient} illustrates five distinct survival curves generated by our survival model, corresponding to five different patients from the test cohort with events equal to one (deceased). Based on the ground truth survival time, patient 1 died after 645 days, patient two after 688 days, patient three after 102 days, patient four after 2,000 days, and patient five after 39 days.

At the time of their respective deaths, the model predicted survival probabilities of 0.42, 0.15, 0.3, 0.05, and 0.5 for patients 1 through 5. These values indicate varying degrees of accuracy in predicting the survival probabilities at the actual time of death, with patient 4 exhibiting the lowest probability and patient 5 the highest. At 500 days, the model's survival probability predictions for patients 1 to 5 were 0.57, 0.2, 0.06, 0.3, and 0.05, respectively. At 1000 days, these probabilities decreased to 0.1, 0.07, 0, 0.18, and 0 for the same patients. At 1500 days, all survival probability predictions reached 0, except for patient 4, whose probability reached 0 at 2000 days. The above findings suggest that the model demonstrates varying performance in predicting survival probabilities for the five patients at different time points. Some predictions align closely with the ground truth survival times, while others exhibit a bit of discrepancy.

\begin{figure}[!h]
\centering
\begin{tikzpicture}  
\begin{axis} 
[domain = 0:100, 
axis on top=false, 
axis x line=middle, 
axis y line=middle, 
xlabel = \fontfamily{qhv}\small Time (Days), 
xlabel near ticks, 
ylabel = \fontfamily{qhv}\small Survival Probabilities,  
ylabel near ticks, 
width=7.0cm, 
legend style={font=\scriptsize, at={(1.025,1)},anchor=north west,legend columns=1}, 
height=5.5cm, 
clip  = true, 
xmin = 0,   
xmax = 3300, 
ymin = 0, 
ymax = 1.05, 
grid=major, 
ytick={0,0.05,0.2,0.4,0.6,0.8,1,1.2}, 
yticklabels={$0$,$0$,$0.2$,$0.4$,$0.6$,$0.8$,$1$}, 
xtick={0,5,500,1000,1500,2000,2500,3000,3500}, 
xticklabels={$0$,$0$,$500$,$1000$,$1500$,$2000$,$2500$,$3000$,$3500$},]

\addplot [Salmon, line width = 0.9, smooth] coordinates {(0,1)(250,0.99)(500,0.98)(1000,0.97)(2000,0.96)(3000,0.95)};
\addlegendentry{Patient 6}

\addplot [TealBlue, line width = 0.9, smooth] coordinates {(0,1)(500,0.95)(1000,0.94)(1500,0.92)(2000,0.9)(2200,0.88)(2500,0.83)(2700,0.78)(3000,0.73)};  
\addlegendentry{Patient 7}

\addplot [Sepia, line width = 0.9, smooth] coordinates {(0,1)(500,0.97)(1000,0.95)(1500,0.93)(2000,0.9)(2200,0.88)(2500,0.85)(2700,0.8)(3000,0.72)};  
\addlegendentry{Patient 8}

\addplot [Thistle, line width = 0.9, smooth] coordinates {(0,1)(500,0.95)(1000,0.93)(1500,0.9)(2000,0.88)(2200,0.83)(2500,0.7)(2700,0.6)(3000,0.4)};  
\addlegendentry{Patient 9}

\addplot [Goldenrod, line width = 0.9, smooth] coordinates {(0,1)(500,0.98)(1000,0.96)(1500,0.94)(2000,0.91)(2200,0.88)(2500,0.79)(2600,0.75)(2800,0.65)(3000,0.55)};  
\addlegendentry{Patient 10}

\end{axis}  
\end{tikzpicture} 
\caption{Survival Probabilities for five patients in the test cohort who had censored events.}
\label{survivalcurvealive}
\end{figure}
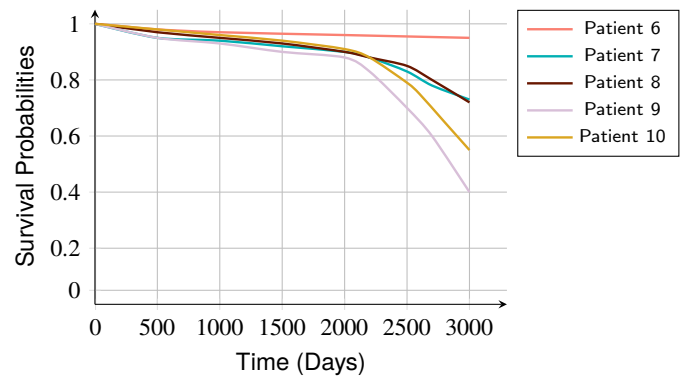

Fig. \ref{survivalcurvealive} illustrates five distinct survival curves generated by our survival model for five different patients from the test cohort with events equal to zero (censored) and censoring time greater than 2000 days. Based on the ground truth survival time, their censoring times are 2473 days for patient 6, 2045 days for patient 7, 2900 days for patient 8, 2600 days for patient 9, and 2298 days for patient 10.

For patient 6, the model indicates a high probability of survival (0.95) at the censoring time of 2473, while patient 7 has a slightly lower survival probability of 0.9 at the censoring time of 2045. Patients 8, 9, and 10 exhibit survival probabilities of 0.75, 0.68, and 0.87 at their censoring times of 2900, 2600, and 2298, respectively. These predictions suggest that patient 6 has the highest likelihood of survival at their censoring time, followed by patients 7 and 10. Conversely, patients 8 and 9 possess relatively lower survival probabilities, with patient 9 exhibiting the lowest probability of survival among the five patients at their respective censoring times.

\subsection{Violin Diagram for Survival Distribution}
\begin{figure}[!ht]
    \vspace{-0.5em}
    \centering
    \vspace{1ex}
    \includegraphics[trim=0.25cm 0.5cm 1.5cm 1.0cm, clip=true, scale=0.625]{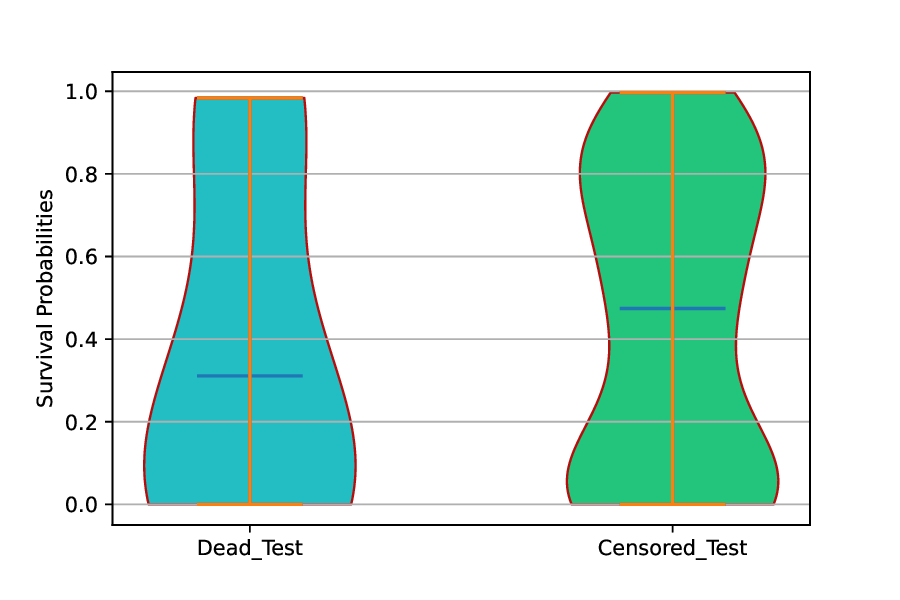}
    \caption{Violin plots for censored \& deceased events in train \& test sets.}
    \label{violplot}
    \vspace{-0.4em}
\end{figure}

Fig. \ref{violplot} presents the violin plot for censored and uncensored subjects in the testing subset, showcasing the survival probability on the vertical axis for Exp8, which emerged as the optimal experimental outcome. As we mentioned in Section \ref{metricseval}, with violin plots, we can comprehend the distribution of survival probabilities predicted by our survival model. 

{Censored\_Test} relates to the patients who did not experience the event in the test subset. Regarding the {Censored\_Test}, we are uncertain about the outcomes at the final time point (whether death occurred or not). Based on the median, it can be inferred that for half of the subjects, a survival probability lower than 0.45 would be predicted, with a higher concentration around 0.1. Conversely, for the remaining half, a survival probability greater than 0.45 would be anticipated, with a greater distribution around 0.8. Given the symmetrical distribution around the median for {Censored\_Test}, the model predicts that half of the censored patients would exhibit high survival probability at the last observation time. In contrast, the other half would demonstrate low survival probability. {Dead\_Test} refers to patients who died within the test subset. This group's ideal output survival probabilities distribution is at zero. The median survival probability predicted by our survival model  is around 0.3. Our survival model accurately predicted near-zero survival probabilities for half of the patients whose predicted probabilities were below the median. The other half of the patients with predicted probabilities higher than the median had distributions mostly near the median. Those nearer to the median had accurate survival predictions but with a small time shift. Those close to 1 are those patients whose survival probabilities were not accurately calculated. Upon analyzing the violin plots of the test subset for both censored and deceased patients, it can be concluded that our proposed multimodal survival model yields satisfactory outcomes that mostly align closely with the actual follow-up times of patients.

\section{Discussion}
The hypotheses underlying our study were twofold. Firstly, we aimed to investigate whether the selective provision of the most relevant clinical variables to the model would enhance the performance evaluation of survival analysis, as opposed to indiscriminately supplying all clinical variables. As evidenced by Table \ref{9experimentstable} in Section \ref{results}, our findings revealed that the most favorable results were obtained in Exp 8, wherein clinical variables were judiciously chosen. In contrast, Exp 3, which involved the inclusion of all clinical variables, yielded a lower C-index (by 0.02) and a reduced AUC (by 0.06). Our second hypothesis posited that multimodal survival analysis would yield superior results when compared to single-modality approaches. In support of this hypothesis, Table \ref{9experimentstable} in Section \ref{results} demonstrates that using single-modality data, such as solely clinical data or CT image features, led to lower performance metrics. In contrast, Exp 3 through 9, which incorporated a combination of clinical data and CT image features, resulted in significantly improved performance outcomes.

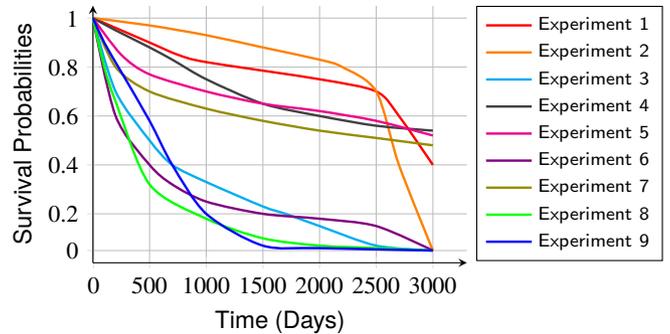
\begin{figure}[!h]
\centering
\begin{tikzpicture}  
\begin{axis} 
[domain = 0:100, 
axis on top=false, 
axis x line=middle, 
axis y line=middle, 
xlabel = \fontfamily{qhv}\small Time (Days), 
xlabel near ticks, 
ylabel = \fontfamily{qhv}\small Survival Probabilities,  
ylabel near ticks, 
width=6.5cm, 
legend style={font=\scriptsize, at={(1.025,1)},anchor=north west,legend columns=1}, 
height=5cm, 
clip  = true, 
xmin = 0,   
xmax = 3300, 
ymin = 0, 
ymax = 1.05, 
grid=major, 
ytick={0,0.05,0.2,0.4,0.6,0.8,1,1.2}, 
yticklabels={$0$,$0$,$0.2$,$0.4$,$0.6$,$0.8$,$1$}, 
xtick={0,5,500,1000,1500,2000,2500,3000,3500}, 
xticklabels={$0$,$0$,$500$,$1000$,$1500$,$2000$,$2500$,$3000$,$3500$},]

\addplot [red, line width = 0.9, smooth] coordinates {(0,1)(500,0.9)(750,0.85)(1000,0.82)(2000,0.75)(2500,0.7)(2700,0.6)(3000,0.4)};  
\addlegendentry{Experiment 1}

\addplot [orange, line width = 0.9, smooth] coordinates {(0,1)(500,0.97)(1000,0.93)(1500,0.88)(2000,0.83)(2200,0.8)(2500,0.7)(2700,0.4)(3000,0.05)};   
\addlegendentry{Experiment 2}

\addplot [cyan, line width = 0.9, smooth] coordinates {(0,1)(200,0.7)(500,0.5)(700,0.4)(1000,0.33)(1500,0.23)(1700,0.2)(2000,0.15)(2500,0.07)(3000,0.05)};  
\addlegendentry{Experiment 3}
    
\addplot [darkgray, line width = 0.9, smooth] coordinates {(0,1)(200,0.95)(500,0.88)(700,0.83)(1000,0.75)(1500,0.65)(2000,0.6)(2500,0.56)(3000,0.54)};  
\addlegendentry{Experiment 4}

\addplot [magenta, line width = 0.9, smooth] coordinates {(0,1)(250,0.85)(500,0.77)(1000,0.7)(1500,0.65)(2000,0.62)(2500,0.58)(3000,0.52)};  
\addlegendentry{Experiment 5}

\addplot [violet, line width = 0.9, smooth] coordinates {(0,1)(200,0.6)(500,0.4)(700,0.32)(1000,0.25)(1500,0.2)(2000,0.18)(2500,0.15)(3000,0.05)};  
\addlegendentry{Experiment 6}

\addplot [olive, line width = 0.9, smooth] coordinates {(0,1)(200,0.8)(500,0.7)(1000,0.63)(1500,0.58)(2000,0.54)(2500,0.51)(3000,0.48)};  
\addlegendentry{Experiment 7}

\addplot [green, line width = 0.9, smooth] coordinates {(0,1)(100,0.8)(200,0.66)(500,0.32)(1000,0.18)(1500,0.1)(2000,0.07)(2500,0.06)(3000,0.05)};  
\addlegendentry{Experiment 8}

\addplot [blue, line width = 0.9, smooth] coordinates {(0,1)(100,0.9)(250,0.78)(500,0.58)(700,0.4)(1000,0.2)(1500,0.07)(2000,0.06)(3000,0.05)};  
\addlegendentry{Experiment 9}

\end{axis}  
\end{tikzpicture} 
\caption{Survival Probabilities from 9 different experiments for one patient.\label{fig:S(t)}}
\label{survivalcurve9exp}
\end{figure}

To demonstrate that the integration of clinical data and CT 
image features results in superior performance compared to using CT image features or clinical data alone; we selected a single patient from the test cohort whose survival curve was incorrectly plotted in Exp 1 and Exp 2, in which both used a single data modality. This patient had an ISUP grade of 4 and a survival duration of 2,000 days. Subsequently, we generated survival curves for this patient from our nine defined experiments as illustrated in Fig. \ref{survivalcurve9exp}. The estimated survival probabilities for the selected patient at the time of death (2,000 days) were approximately 0.77 and 0.82 for Exp 1 and Exp 2, respectively. In contrast, the survival probabilities at the time of death for Exp 3 through 9 were as follows 0.18 for Exp 3, 0.6 for Exp 4, 0.61 for Exp 5, 0.19 for Exp 6, 0.55 for Exp 7, 0.05 for Exp 8, and 0 for Exp 9. This result demonstrates that multimodal data can yield superior results compared to single-modality experiments.

In the context of our study, we sought to draw comparisons with other studies that employed radiological images and clinical variables as inputs for their survival models. A summary of these methodologies can be found in Section \ref{relatedwork}. Table \ref{comparisontable} presents a comparison between our approach and previous studies, focusing on the C-index and AUC metrics. Our method outperforms the others in terms of both C-index and AUC, as demonstrated in Table \ref{comparisontable}. Our methodology, utilizing 17 clinical variables, yielded the highest C-index and AUC values, demonstrating its superior performance. As indicated in the second row of the table \ref{comparisontable} for Exp4, our approach's effectiveness remains evident even when only four clinical variables are employed. The C-index and AUC values in Exp4 scenario continue to surpass those of alternative methods, despite the constrained number of clinical variables utilized. Additionally, it is worth noting that none of the aforementioned studies provided a methodology capable of generating non-proportional individualized survival curves for distinct patients. Furthermore, these studies relied on traditional methodologies that were susceptible to proportionality issues. In contrast, our approach not only yielded superior performance in terms of C-index and AUC but also addressed the limitations inherent in previous studies.

\begin{table}[h]
\centering
\caption{Comparison of this study results with previous related studies.}
\small
\begin{tabular}{|p{2.5cm}|p{2.45cm}|p{1.1cm}|p{0.5cm}|}
\hline
\textbf{Studies} & \textbf{Number of Clinical Variables} & \textbf{C-index} & \textbf{AUC}\\
\hline\hline
\rowcolor{lemon}\textbf{Our Method} & 17 (Exp8) & 0.84 & 0.8\\
\hline
\rowcolor{lemon}\textbf{Our Method} & 4 (Exp4) & 0.79 & 0.76\\
\hline
\textbf{\citet{paper1compare}} & 2 (Age, TNM Stage) & - & 0.76\\
\hline
\textbf{\citet{DeepMMSA:surv}} & 5 (Age, Histology, TNM Stage, Overall Stage, Gender) & 0.65 & -\\
\hline
\textbf{\citet{gastric:surv}} & 3 (Tumor Size,\newline Tumor Localization, TNM Stage) & 0.78 & -\\ 
\hline
\textbf{\citet{nasopha:surv}} & 3 (Age, LDH, \newline Pre-EBV DNA) & 0.78 & -\\
\hline
\end{tabular}
\label{comparisontable}
\end{table}

In addition to the benefits of our method, our study has a number of limitations. Firstly, for Experiment 8, which achieved the highest C-index and AUC, 17 clinical variables were employed during the training process. In order to generate survival predictions for a new patient, it is essential to obtain all 17 clinical variables to ensure the accuracy of the survival estimation. Secondly, precise feature extraction necessitates not only whole abdomen images but also segmentation annotations of the target organ and associated tumors. Thirdly, to generalize this study's findings to other types of cancer, it is essential to pinpoint a clinical variable comparable to the ISUP grade, enabling tumor classification in relation to survival estimation.

In future research, we aim to explore the feasibility of integrating RCC ISUP grade classification and survival prediction within a unified training framework, eliminating the need for separate tumor grading. Furthermore, we intend to investigate innovative approaches for feature extraction that circumvent the necessity for organ and tumor annotations, thereby enhancing the applicability and efficiency of the proposed methodology.

\section{Conclusion}
This study presents a novel multimodal AI-based framework for predicting individualized survival probabilities of patients with renal cell carcinoma. The proposed framework utilizes CT imaging and clinical data as inputs. We demonstrated that relevant features for survival estimation could be extracted from CT scans and combined with clinical data to improve performance. Our proposed framework can generate personalized, non-linear, and non-proportional survival probability curves for different patients, achieving higher accuracy and outperforming previously published methods. We showed that using a multimodal  strategy for survival analysis leads to higher accuracy than a single-modality approach. Moreover, we presented that carefully selecting significant clinical factors as inputs to the survival model can further enhance the performance of survival prediction. This study lays the path for enhanced clinical decision-making for renal cell carcinoma patients, allowing for more precise and individualized therapy options based on the combination of radiological imaging and clinical data. Future research in this field may build upon these findings, resulting in even more complex and reliable survival prediction models.

\section{Acknowledgement}
The authors acknowledge the CIRCLE grant no. 287112 and the Health South-East Trust grant no. 2023069 for funding this study. We thank H{\aa}vard Kvamme, a previous Ph.D. student at the University of Oslo, for his invaluable guidance in effectively utilizing the pycox library he created.

\bibliographystyle{model2-names.bst}
\bibliography{cas-refs}

\end{document}